\documentclass[a4paper, 10pt, conference]{mhs}      

\IEEEoverridecommandlockouts                              

\usepackage{graphics} 
\usepackage{epsfig} 
\usepackage{mathptmx} 
\usepackage{times} 
\usepackage{amsmath} 
\usepackage{amssymb}  
\usepackage{here}
\usepackage{threeparttable}
\usepackage{caption}

\title{\Large \bf
 Voice control interface for surgical robot assistants}

\author{{\large Ana Davila$^{1}$, Jacinto Colan$^{2}$ and Yasuhisa Hasegawa$^{1}$}\\
{\normalsize $^{1}$ Institutes of Innovation for Future Society, Nagoya University, Aichi, Nagoya, Japan}\\{\normalsize $^{2}$ Dept. of Micro-Nano Mechanical Science and Engineering, Nagoya University, Aichi, Nagoya, Japan}
\\\\}

\begin{document}
    
    \maketitle
    \thispagestyle{empty}
    \pagestyle{empty}

    \begin{abstract}
    Traditional control interfaces for robotic-assisted minimally invasive surgery impose a significant cognitive load on surgeons.
     To improve surgical efficiency, surgeon-robot collaboration capabilities, and reduce surgeon burden, we present a novel voice control interface for surgical robotic assistants. Our system integrates Whisper, state-of-the-art speech recognition, within the ROS framework to enable real-time interpretation and execution of voice commands for surgical manipulator control. The proposed system consists of a speech recognition module, an action mapping module, and a robot control module. Experimental results demonstrate the system's high accuracy and inference speed, and demonstrates its feasibility for surgical applications in a tissue triangulation task. Future work will focus on further improving its robustness and clinical applicability.
    \end{abstract}

    
    \section{Introduction}
    Surgical robotics has significantly advanced surgical capabilities, offering enhanced precision, dexterity, and minimally invasive procedures. However, traditional control interfaces, which rely primarily on joysticks and graphical user interfaces, often present a cognitive burden on surgeons, particularly in the high-stress environment of the operating room. These interfaces can be complex, require significant attention and potentially hinder surgeon focus on critical aspects of the procedure.
    
    Autonomous surgical systems offer the potential to alleviate the control burden on surgeons by making context-aware decisions. For instance, they could anticipate surgical phase transitions or gestures and activate predefined assistance \cite{yamada2024multimodal, chen2022realtime, staub2011implementation}, or track dynamic regions of interest to autonomously manipulate endoscopes \cite{fozilov2023endoscope}. However, current autonomous systems still struggle to fully understand the complexities of surgical environments and tasks. As a result, direct human control remains indispensable for ensuring patient safety and achieving optimal surgical outcomes.
    
    In parallel, surgical teams rely heavily on verbal communication to coordinate their actions and make critical decisions. This suggests that voice control could offer a natural and intuitive interface for interacting with surgical robotic assistants. Previous research has explored voice-based control for surgical robotics and was found to be feasible \cite{kubben2019feasibility}, but challenges such as limited vocabulary recognition, latency issues, and the demanding acoustic environment of the operating room have hindered its widespread adoption \cite{schulte2020automatic}.
    
    To address these challenges, we propose a novel voice control interface for surgical robotic assistants. Our system incorporates recent advances in speech recognition technology to accurately capture and interpret spoken commands. By mapping these commands to predefined robot actions, we aim to create a more intuitive and efficient interaction framework. The resulting robot motion commands are transmitted to a dexterous robotic manipulator to perform surgical assistance tasks.

    \begin{figure}[t]
        \centering
        \includegraphics[width=\columnwidth]{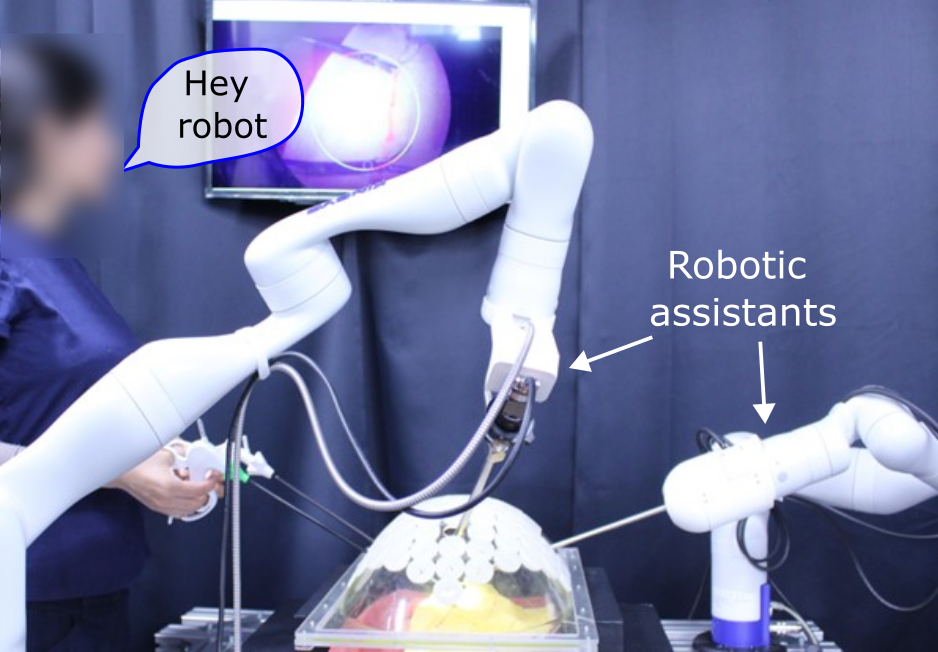}
        \caption{Voice commands can be used for commanding robotic surgical assistants.}
        \label{fig:1}
    \end{figure}
    
    \section{Related works}
    
    Voice control has been explored as an alternative to traditional input methods for surgical robotics, with the aim of improving efficiency and reducing the cognitive load on surgeons. Early studies, such as those of Allaf et al. \cite{allaf1998laparoscopic}, compared voice and foot pedal interfaces for controlling the AESOP robot, demonstrating the potential of voice commands to reduce the cognitive load on surgeons. Nathan et al. \cite{nathan2006voice} further investigated the use of the AESOP system in endoscopic surgeries, highlighting its effectiveness in improving surgical precision and reducing operation time.
    
    El-Shallaly et al. \cite{elshallaly2005voice} examined the impact of voice recognition interfaces (VRI) on enabling the surgeon to perform and control the light source, camera, and insufflator during laparoscopic cholecystectomy, finding significant improvements in operating time and staff efficiency. Zinchenko et al. \cite{zinchenko2016study} conducted a study on intentional speech recognition control for surgical robotic endoscopes to address the ambiguity of robotic motion for each speech command. Their results emphasize the importance of accurate and reliable voice command recognition in complex surgical environments.
    
    He et al. \cite{he2021design} designed a voice-based control system using a commercial speech recognition interface for a nasal endoscopic surgical robot in simulation environments. The proposed method analyzes eight motion directions of the endoscopic image and obtains the corresponding motion speed of the endoscope tip under RCM constraints.
    
    Previous works have relied on offline speech recognition toolkits. Among these modules, Vosk and Kaldi \cite{povey2011kaldi} have provided a robust foundation for implementing voice control in various robotic applications. In addition to offline modules, recent studies on cloud-based speech recognition systems have shown the potential of cloud technologies to enhance the performance and scalability of voice-controlled robotic systems \cite{deuerlein2021human}. Recent advances have focused on comparing different control interfaces. Yang et al. \cite{yang2020novel} compared a novel foot interface with voice control for robotic endoscope holders using Google Cloud Speech. The results showed that the foot interface performed better than the voice system in the average completion time and error rate, with a lower cognitive burden, providing insights into the advantages and limitations of each method.
    
    Elazzazi et al. \cite{elazzazi2022natural} and Paul et al. \cite{paul2024evaluation} explored natural language interfaces for autonomous camera control systems on the da Vinci Surgical Robot. In \cite{elazzazi2022natural}, an offline speech recognition module (Vosk) was compared to a cloud speech recognition module (Alexa), obtaining better performance with the cloud-based one in virtual environments. In \cite{paul2024evaluation}, human studies showed a preference for voice control, using Kaldi, compared to manual control using a joystick.
    
    Recent work in other fields has used deep learning models for speech recognition. Domínguez-Vidal and Sanfeliu \cite{dominguez2024voice} investigated the recognition of voice commands for collaborative object transportation tasks, using convolutional neural networks (CNNs) to process spectrograms of voice commands and recognize the speech command.
    
    Although these studies provide valuable insights into voice control for surgical robotics, there is still a need for further research to address challenges such as robustness in noisy environments and integration with existing surgical workflows. Furthermore, the potential of combining voice control with other advanced technologies, such as artificial intelligence and machine learning, remains largely unexplored and could offer significant improvements in the efficiency and effectiveness of surgical robotic systems.

    \section{Methodology}
\begin{figure}[t]
        \centering
        \includegraphics[width=\columnwidth]{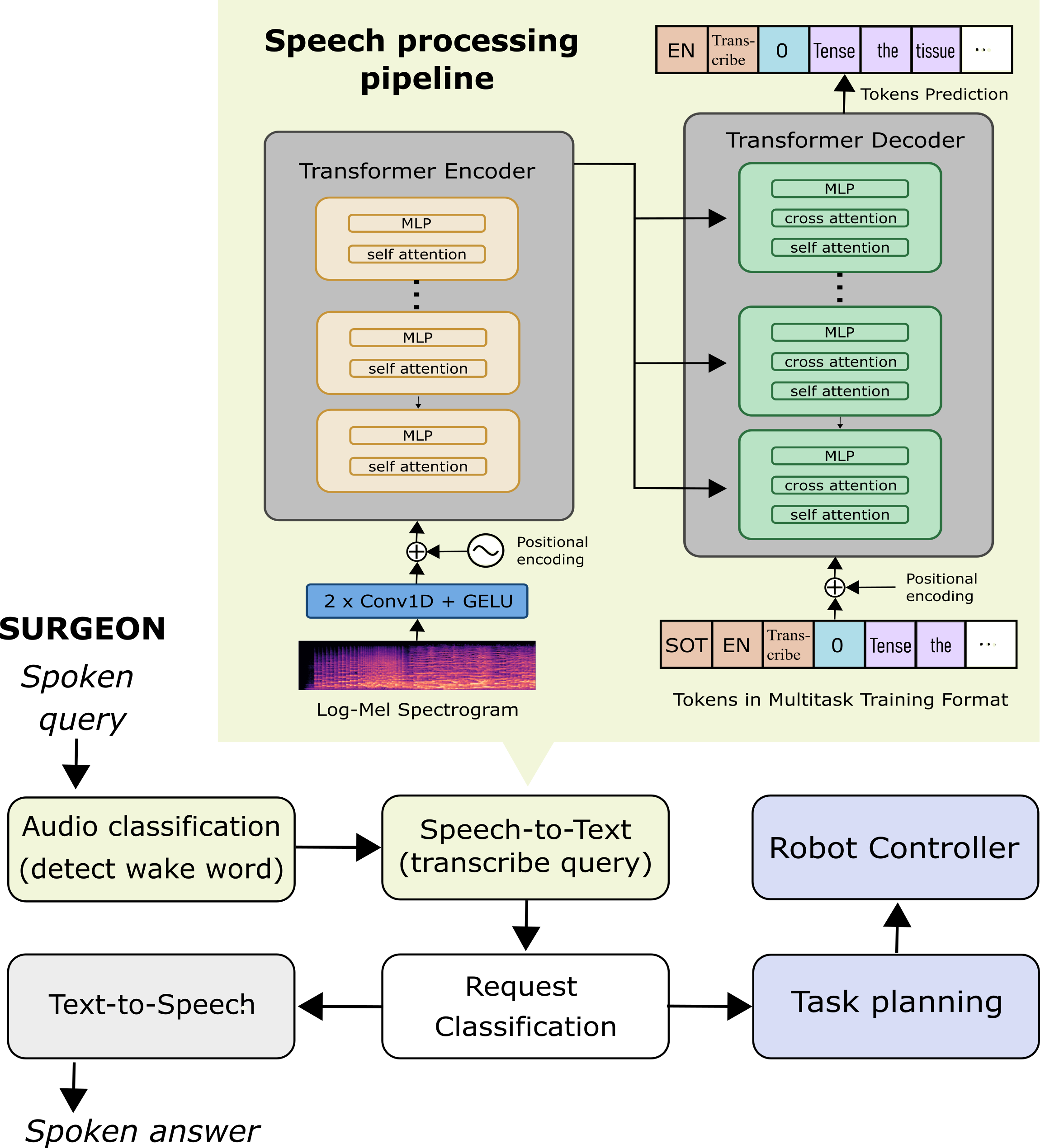}
        \caption{Overview of the proposed voice control.}
        \label{fig:2}
    \end{figure}

    The proposed system consists of a speech recognition module (SRM), a mapping module (MM) and a robotic manipulator equipped with a robotic surgical tool.  A block diagram of the functionality of the proposed system is shown in Fig.~\ref{fig:2}.
    
    \subsection{Speech recognition unit}
    The Speech Recognition Module (SRM) is responsible for accurately capturing and interpreting spoken commands from the surgeon. The SRM has three main functions: recording the voice input, performing speech recognition, and generating the transcript of the requested command. As the core for the SRM, we utilize Whisper \cite{radford2023robust}, an advanced speech recognition model developed by OpenAI.
    
    Whisper is a state-of-the-art automatic speech recognition (ASR) system that leverages a large-scale, multilingual, and multitask supervised dataset collected from the web. This extensive dataset enables Whisper to achieve high robustness and accuracy, even in challenging acoustic environments. The model is designed to handle various speech processing tasks, including multilingual speech recognition, speech translation, spoken language identification, and voice activity detection.
    
    The Whisper architecture is based on a Transformer sequence-to-sequence model, which is trained on diverse audio data. The input audio is split into 30-second chunks, converted into a log-Mel spectrogram, and then passed to an encoder. The decoder predicts the corresponding text caption, intermixed with special tokens that direct the model to perform specific tasks such as language identification and phrase-level timestamps. One of the key advantages of Whisper is its ability to handle background noise and accents effectively, making it suitable for the noisy and dynamic environment of an operating room. In addition, Whisper supports transcription in multiple languages and translation from those languages into English, providing flexibility in multilingual surgical settings.

    The SRM workflow begins with the recording of the surgeon’s voice input using a high-quality microphone embedded into a bluetooth headset. The recorded audio is then preprocessed with noise reduction and filtering to enhance the signal quality and reduce noise. Next, the preprocessed audio is fed into the Whisper model, which performs the speech recognition task. The model processes the audio in real-time, generating a transcript of the spoken commands. This transcript is then passed on to the mapping algorithm (MA), which interprets the commands and translates them into specific actions for the robotic manipulator.
        
    \subsection{Mapping module}
    The mapping module is responsible for interpreting transcribed voice commands and mapping them to specific robotic actions. In the context of a robotic system assisting with tissue manipulation tasks, we have defined seven distinct commands to control robot motion.
    
    \begin{itemize} 
        \item \textbf{“hey robot”}: Activates robot assistance mode. This command initializes the system, making it ready to receive further instructions from the surgeon. 
        \item \textbf{“tense”}: Generates a sequence of actions comprising reach, grasp (close gripper), and pull. This command is used to create tension in the tissue by pulling it in a specific direction. 
        \item \textbf{“release”}: Opens the gripper to release the tissue. 
        \item \textbf{“pull more”}: Increases tension by pulling further in the pulling direction. 
        \item \textbf{“pull less”}: Reduces tension by retracting the forceps in the pulling direction.  
        \item \textbf{“stop”}: Disables robot motion for safety. This command is used in emergency situations to immediately halt all robot movements.
        \item \textbf{“terminate”}: Deactivates robot assistance mode. This command stops the robot from receiving further instructions and returns it to a standby state.
    \end{itemize}
    
    The mapping module computes the Word Error Rate (WER) between the transcribed command and each of the predefined commands. WER is a common metric used in speech recognition to measure the accuracy of the transcription. It is calculated as the sum of the substitutions, deletions, and insertions required to transform the transcribed text into the reference text, divided by the number of words in the reference text. The computed WER for each predefined command is compared against a predefined threshold. This threshold is set to ensure that only commands with a high degree of accuracy are considered for execution. If the WER for a particular command is below the threshold, it is deemed a match. The command with the lowest WER that meets the threshold criteria is selected as the intended command. This ensures that the most accurate interpretation of the spoken command is chosen. The selected command is then sent to the robot controller for execution. 
    
    \subsection{Robot control}
    The robot controller is responsible for translating the commands provided by the Matching Algorithm (MA) into specific actions for the robotic manipulator. The Robotic Operating System (ROS) framework is utilized to provide modularity and interconnection between various modules, ensuring seamless integration and communication. The Speech Recognition Module (SRM) and the Matching Algorithm (MA) are integrated into a single ROS node, while the robot control is implemented as an independent ROS node. 

    Communication between the SRM/MA node and the robot control node is achieved through the use of ROS services. When the MA selects an appropriate action based on the transcribed voice command, it generates a service request. This request embeds the action command as a string variable, which is then sent to the robot controller. The robot controller receives the action command, verifies that the action is achievable, and confirms the reception of the command. This verification step ensures that the robot can safely and effectively perform the requested action. Internally, the robot controller uses the received command to feed an action server. The action server is responsible for planning the constrained motion, generating trajectories, and executing the robot motion with respect to an RCM constrained \cite{davila2024realtime}. 
    
    \section{Experimental validation}
    We evaluated the performance of the speech recognition module and demonstrated its feasibility in a robot-assisted tissue manipulation task. 
    
    \subsection{Experimental setup}
\begin{figure}[t]
        \centering
        \includegraphics[width=\columnwidth]{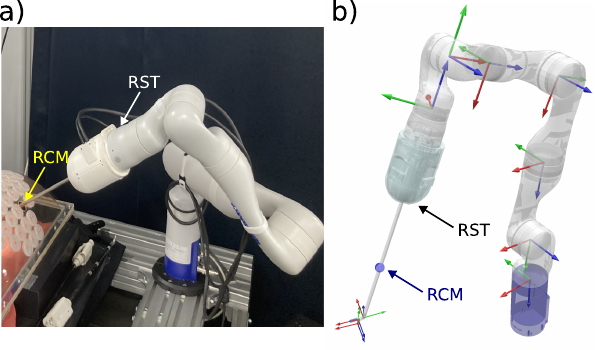}
        \caption{Robotic surgical assistant. a. Robotic manipulator comprising a 7-DOF manipulator and a 3-DOF robotic surgical tool (RST). b. Kinematic description}
        \label{fig:3}
    \end{figure}

    Our experimental setup comprises a 7-DoF robotic manipulator (Gen3, Kinova) equipped with a 3-DoF robotic surgical tool (OpenRST\cite{colan23openrst}). This setup serves as the surgical robotic assistant, capable of performing precise and controlled tissue manipulation tasks. The robotic manipulator is shown in Fig.~\ref{fig:3}. The voice interface utilizes a Bluetooth headset (Flex, Beats) to capture surgeon voice commands. The headset was chosen for its high-quality audio capture and noise-canceling features. The software implementation of our approach was carried out using the PyTorch 2.0 framework, on a workstation equipped with an AMD Ryzen Threadripper PRO 3995WX 2.7GHz processor, 256 GB RAM, and an NVIDIA GeForce RTX A6000 GPU.

    \subsection{Speech recognition accuracy}
\begin{figure}[t]
        \centering
        \includegraphics[width=0.8\columnwidth]{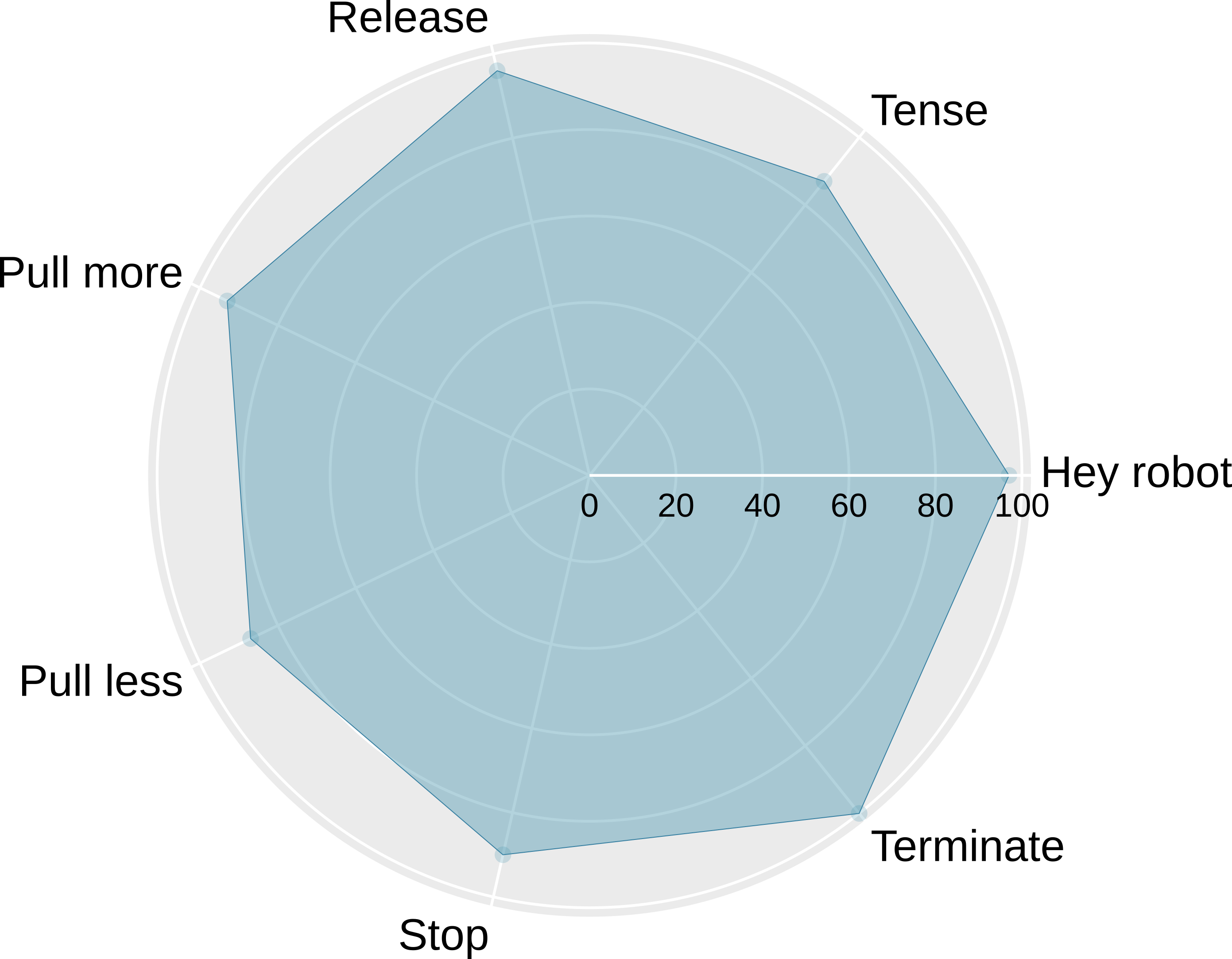}
        \caption{Results for Speech command recognition}
        \label{fig:4}
    \end{figure}

    We evaluated the recognition accuracy for the seven commands defined in Section 3.2. A voice command is considered correctly recognized if it matches the predicted action command. Two subjects, both non-native English speakers, repeated each of the voice commands 30 times. The recognition accuracy is shown in Fig.~\ref{fig:4}, demonstrating that most of the commands exhibit high recognition performance.

    \subsection{Inference time}
    Inference time is measured as the duration between the completion of the voice request by the user and the initiation of the robot’s action. Table~\ref{table:1} summarizes the inference times recorded for the seven voice commands, each repeated 30 times. The average inference time is approximately 1.7 seconds, which is generally sufficient for high-level and general voice requests.

     \begin{table}[hbtp]
        \caption{Inference time (s) for voice command recognition.}
    
        \label{table:1}
        \centering
        \normalsize
        \begin{tabular}{l c }
        \cline{1-2}
        Voice command     & Average time \\ \hline
        hey robot                 &  1.97\\ 
        tense                 &  1.14 \\
        release                 &  1.42\\
        pull more                 &  2.39\\
        pull less                 &  2.50\\
        stop                 &  1.29\\
        terminate                 &  1.52\\  \hline
        \end{tabular}
    \end{table}
    

    \subsection{Demonstration in a tissue triangulation task}
\begin{figure}[t]
        \centering
        \includegraphics[width=\columnwidth]{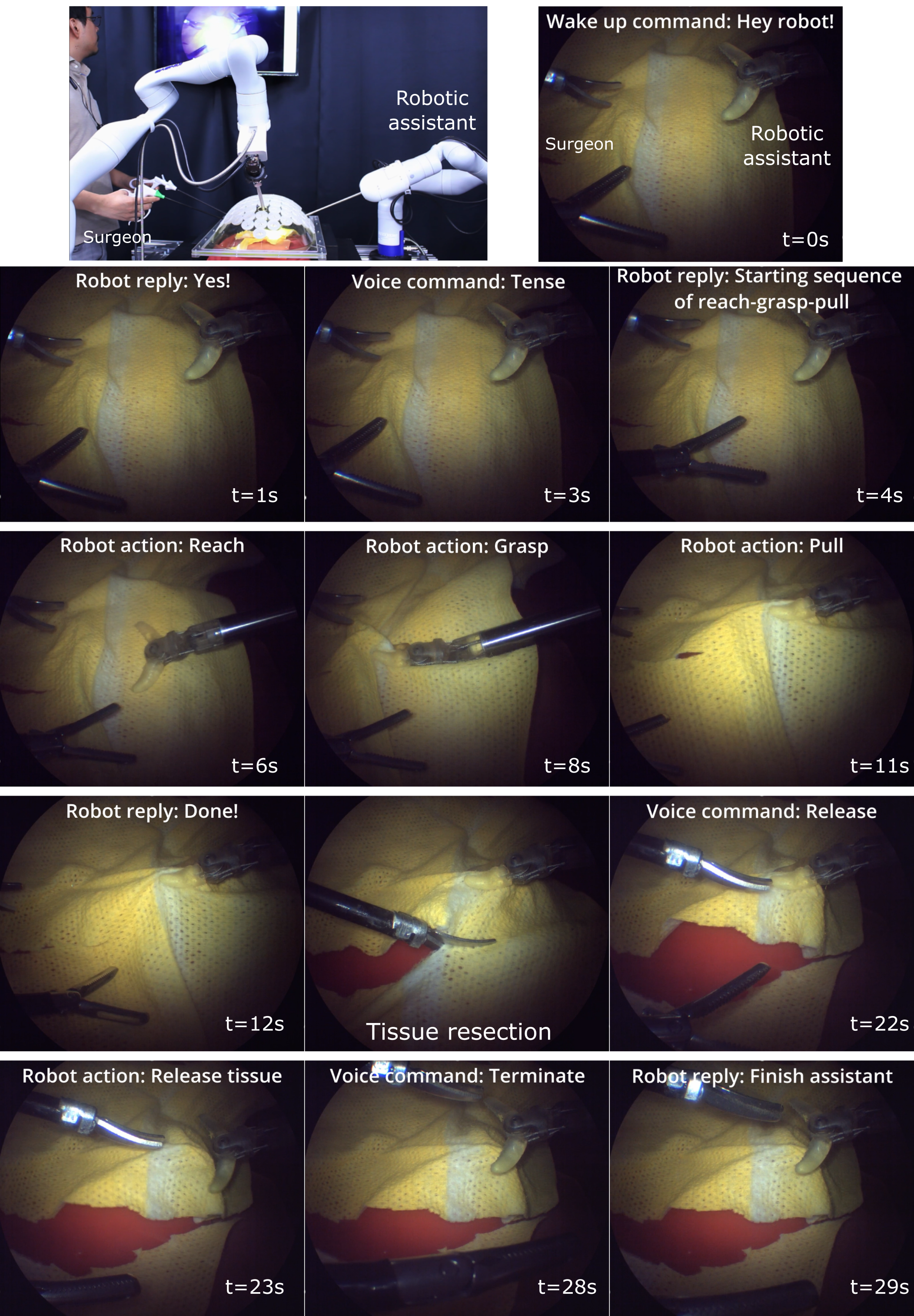}
        \caption{Snapshots of control of a robotic assistant for tissue manipulation.}
        \label{fig:5}
    \end{figure}

    We demonstrated the feasibility of the proposed framework through a tissue triangulation task, as described in \cite{liu2024latent}. In this demonstration, a subject manipulated conventional surgical tools, while the robotic system operated the multi-DoF robotic surgical tool. The operator used voice commands to activate robot actions, triangulate the dummy tissue, and perform a tissue resection task. Snapshots of task performance are shown in Fig.~\ref{fig:5}.
    
    During the task, the operator issued voice commands such as “hey robot” to activate the robot assistance mode, “tense” to generate a sequence of actions comprising reach, grasp, and pull to create tension in the tissue, and “release” to open the gripper and release the tissue. Commands like “pull more” and “pull less” were used to adjust the tension by increasing or decreasing the pulling force, respectively. The command “terminate” was used to deactivate the robot assistance mode, and “stop” was available to immediately halt all robot movements for safety.
   
    It should be noted that further performance improvements could be achieved by fine-tuning the pre-trained model \cite{davila2024comparison}. By using voice demonstrations, the speech recognition system can be trained to distinguish specific voice patterns from surgeons who will be using the proposed framework. This personalized training approach can enhance the accuracy and reliability of the speech recognition module, making it more robust to variations in speech patterns and accents.
    
    \section{Conclusions}
    The proposed voice-controlled robotic assistance framework for surgical tasks shows significant potential in enhancing surgical precision and efficiency. Through the integration of advanced speech recognition technology, specifically the Whisper model, the system effectively interprets and executes voice commands in real-time. Experimental validation, including a tissue triangulation task, showcased high recognition accuracy and reliable performance of the proposed system. The modular architecture, which leverages the ROS framework, ensures flexibility and scalability, making the system adaptable to various surgical scenarios. Future improvements, such as personalized training and advanced machine learning techniques, promise to further enhance the robustness and applicability of the system, aiming for more intuitive and efficient surgical robotic assistance.

    \addtolength{\textheight}{-10cm}   
    

    \section*{ACKNOWLEDGMENT}
    This work was supported in part by the Japan Science and Technology Agency (JST) CREST under Grant JPMJCR20D5, and in part by the Japan Society for the Promotion of Science (JSPS) Grants-in-Aid for Scientific Research (KAKENHI) under Grant 22K14221.
    

    

\end{document}